\documentclass[11pt,a4paper]{article}
\usepackage[hyperref]{emnlp2018}
\usepackage{times}
\usepackage{latexsym}

\usepackage{url}
\usepackage{amsmath}
\usepackage{amssymb}
\usepackage{amsfonts}
\usepackage{graphicx}
\usepackage{color}
\usepackage{multicol}
\usepackage{multirow}
\usepackage{makecell}
\aclfinalcopy
\usepackage{algorithm}
\usepackage[noend]{algpseudocode}

\usepackage{multirow}
\usepackage{amsmath}
\usepackage{capt-of}
\usepackage{tabularx}
\usepackage[caption=false]{subfig}
\usepackage{epsfig}
\usepackage{amssymb}
\usepackage{amsfonts}
\usepackage{booktabs}
\usepackage{scalerel}
\usepackage[inline]{enumitem}
\usepackage{listings}
\usepackage{varwidth}
\usepackage[export]{adjustbox}
\usepackage{tikz}
\usetikzlibrary{tikzmark}
\usepackage{todonotes}
\usepackage{cleveref}

\usepackage{stmaryrd}
\usepackage{bbm}

\usepackage{algorithm}
\usepackage[noend]{algpseudocode}

\definecolor{deepblue}{rgb}{0,0,0.5}
\definecolor{officeblue}{RGB}{0,102,204}
\definecolor{deepred}{rgb}{0.6,0,0}
\definecolor{deepgreen}{rgb}{0,0.5,0}
\definecolor{mybrickred}{RGB}{182,50,28}

\definecolor{fillcolor}{RGB}{216,217,252}

\algnewcommand\algorithmicrequireb{{\hspace{0.95cm}}}
\algnewcommand\INPTDESCB{\item[\algorithmicrequireb]}

\algnewcommand\algorithmicfuncdesc{\textbf{Function:}}
\algnewcommand\FUNCDESC{\item[\algorithmicfuncdesc]}
\algnewcommand\algorithmicfuncdescb{{\hspace{0.86cm}}}
\algnewcommand\FUNCDESCB{\item[\algorithmicfuncdescb]}
\algnewcommand{\algorithmicgoto}{\textbf{goto}}
\algnewcommand{\Goto}[1]{\algorithmicgoto~\ref{#1}}

\newcommand\AlgComment[1]{\color{mybrickred}{$\triangleright$ \textit{#1}}}

\lstdefinestyle{ifttt}{
	language=Python,
	otherkeywords={Trigger,Action,-,IF,THEN},             
	keywordstyle=\bfseries\color{deepblue},
	emph={MyClass,__init__},          
	emphstyle=\color{deepred},    
	showstringspaces=false,
	breaklines=true,
	escapeinside=||,
	columns=fullflexible,
	basicstyle=\fontfamily{cmtt}\small,
	belowskip=-\baselineskip,
	aboveskip=-0.7\baselineskip
}

\lstdefinestyle{django}{
	language=Python,
	otherkeywords={self},             
	keywordstyle=\bfseries\color{deepblue},
	emph={MyClass,__init__},          
	emphstyle=\color{deepred},    
	showstringspaces=false,
	breaklines=true,
	escapeinside=||,
	columns=fullflexible,
	basicstyle=\fontfamily{cmtt}\small,
	belowskip=-\baselineskip,
	aboveskip=-0.7\baselineskip
}

\lstdefinestyle{pythoncode}{
	language=Python,
	otherkeywords={self},             
	keywordstyle=\bfseries\color{deepblue},
	emph={MyClass,__init__},          
	emphstyle=\color{deepred},    
	showstringspaces=false,
	breaklines=true,
	escapeinside=||,
	columns=fullflexible,
}

\title{Semi-Supervised Learning for Neural Keyphrase Generation}

\author{Hai Ye\thanks{\ \ Work was done while visiting Northeastern University.} \and Lu Wang \\
  College of Computer and Information Science \\
  Northeastern University \\
  Boston, MA 02115 \\
  {\tt hye.me@outlook.com \  luwang@ccs.neu.edu }
}
\date{}

\begin{document}
\maketitle
\begin{abstract}
We study the problem of generating keyphrases that summarize the key points for a given document. While sequence-to-sequence (seq2seq) models have achieved remarkable performance on this task \cite{DeepKey}, model training often relies on large amounts of labeled data, which is only applicable to resource-rich domains. In this paper, we propose semi-supervised keyphrase generation methods by leveraging both labeled data and large-scale unlabeled samples for learning. Two strategies are proposed. First, unlabeled documents are first tagged with synthetic keyphrases obtained from unsupervised keyphrase \emph{extraction} methods or a self-learning algorithm, and then combined with labeled samples for training. Furthermore, we investigate a multi-task learning framework to jointly learn to generate keyphrases as well as the titles of the articles. 
Experimental results show that our semi-supervised learning-based methods outperform a state-of-the-art model trained with labeled data only. 
\end{abstract}

\section{Introduction}
\begin{figure}
\setlength{\abovecaptionskip}{-0.1cm}
\setlength{\belowcaptionskip}{-0.5cm}
\centering\includegraphics[width = \columnwidth]{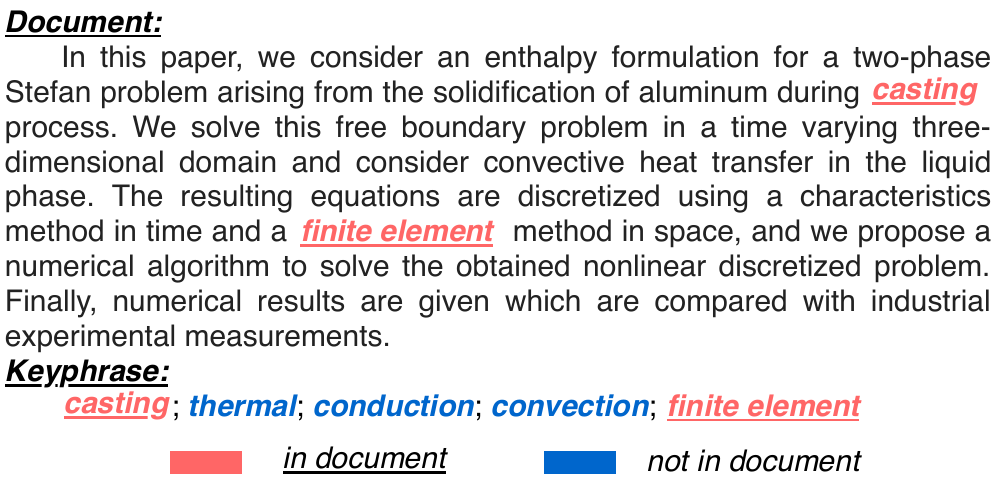}\caption{{Sample document with labeled keyphrases.}}
\label{example}
\end{figure}

Keyphrase extraction concerns the task of selecting a set of phrases from a document that can indicate the main ideas expressed in the input \cite{DBLP:journals/ir/Turney00, DBLP:conf/acl/HasanN14}. 
It is an essential task for document understanding because accurate identification of keyphrases can be beneficial for a wide range of downstreaming natural language processing and information retrieval applications. For instance, keyphrases can be leveraged to improve text summarization systems~\cite{DBLP:journals/wias/ZhangZM04,liu2009unsupervised,wang-cardie:2013:ACL2013}, facilitate sentiment analysis and opinion mining~\cite{wilson2005recognizing,DBLP:conf/ijcnlp/Berend11}, and help with document clustering~\cite{hammouda2005corephrase}. 
Though relatively easy to implement, extract-based approaches fail to generate keyphrases that do not appear in the source document, which are frequently produced by human annotators as shown in Figure \ref{example}. 
Recently, \citet{DeepKey} propose to use a sequence-to-sequence model \cite{Seq2Seq} with copying mechanism for \emph{keyphrase generation}, which is able to produce phrases that are not in the input documents. 

While seq2seq model demonstrates good performance on keyphrase generation \cite{DeepKey}, it heavily relies on massive amounts of labeled data for model training, which is often unavailable for new domains. 
%
To overcome this drawback, in this work, \emph{we investigate semi-supervised learning for keyphrase generation, by leveraging abundant unlabeled documents along with limited labeled data}. 
Intuitively, additional documents, though unlabeled, can provide useful knowledge on universal linguistic features and discourse structure, such as context information for keyphrases and that keyphrases are likely to be noun phrases or main verbs. 
Furthermore, learning with unlabeled data can also mitigate the overfitting problem, which is often caused by small size of labeled training data, and thus improve model generalizability and enhance keyphrase generation performance on unseen data.

Concretely, two major approaches are proposed for leveraging unlabeled data. For the first method, unlabeled documents are first tagged with synthetic keyphrases, then mixed with labeled data for model pre-training. Synthetic keyphrases are acquired through existing unsupervised keyphrase \textit{extraction} methods (e.g., TF-IDF or TextRank \cite{TextRank}) or a self-learning algorithm. The pre-trained model will further be fine-tuned on the labeled data only. 
For the second approach, we propose a multi-task learning (MTL) framework\footnote{We use ``semi-supervised learning" as a broad term to refer to the two methods proposed in this paper.} 
by jointly learning the main task of keyphrase generation based on labeled samples, and an auxiliary task of title generation \cite{DBLP:conf/emnlp/RushCW15} on unlabeled documents. Here one encoder is shared among the two tasks. 
Importantly, we test our proposed methods on a seq2seq framework, however, we believe they can be easily adapted to other encoder-decoder-based systems.

Extensive experiments are conducted in scientific paper domain. Results on five different datasets show that all of our semi-supervised learning-based models can uniformly significantly outperform a state-of-the-art model~\cite{DeepKey} as well as several competitive unsupervised and supervised keyphrase extraction algorithms based on F$_1$ and recall scores. We further carry out a cross-domain study on generating keyphrases for news articles, where our models yield better F$_1$ than a model trained on labeled data only. Finally, we also show that training with unlabeled samples can further produce performance gain even when a large amount of labeled data is available.

\section{Related Work}

\paragraph{Keyphrase Extraction and Generation.}
Early work mostly focuses on the keyphrase extraction task, and a two-step strategy is typically designed. Specifically, a large pool of candidate phrases are first extracted according to pre-defined syntactic templates~\cite{TextRank,SingleRank,DBLP:conf/naacl/LiuPLL09,DBLP:conf/conll/LiuCZS11} or their estimated importance scores~\cite{DBLP:conf/emnlp/Hulth03}. In the second step, re-ranking is applied to yield the final keyphrases, based on supervised learning~\cite{DBLP:conf/ijcai/FrankPWGN99,KEA,DBLP:conf/emnlp/Hulth03,DBLP:conf/semeval/LopezR10,DBLP:conf/mwe/KimK09}, unsupervised graph algorithms~\cite{TextRank,SingleRank,DBLP:conf/ijcnlp/BougouinBD13}, or topic modelings~\cite{DBLP:conf/emnlp/LiuLZS09,DBLP:conf/emnlp/LiuHZS10}. 
Keyphrase generation is studied in more recent work. For instance, ~\citet{DBLP:conf/conll/LiuCZS11} propose to use statistic machine translation model to learn word-alignments between documents and keyphrases, enabling the model to generate keyphrases which do not appear in the input. 
\citet{DeepKey} train seq2seq-based generation models \cite{Seq2Seq} on large-scale labeled corpora, which may not be applicable to a new domain with minimal human labels.
\paragraph{Neural Semi-supervised Learning.}
As mentioned above, though significant success has been achieved by seq2seq model in many NLP tasks~\cite{DBLP:conf/emnlp/LuongPM15,pointer,DBLP:conf/acl/DongL16,DBLP:conf/naacl/WangL16,Hai}, they often rely on large amounts of labeled data, which are expensive to get. 
In order to mitigate the problem, semi-supervised learning has been investigated to incorporate unlabeled data for modeling training~\cite{DBLP:conf/nips/DaiL15,DBLP:conf/emnlp/RamachandranLL17}. For example, neural machine translation community has studied the usage of source-side or target-side monolingual data to improve translation quality~\cite{DBLP:journals/corr/GulcehreFXCBLBS15}, where generating synthetic data~\cite{NMT:SennrichHB16,NMT:Zhang}, multi-task learning~\cite{NMT:Zhang}, and autoencoder-based methods~\cite{DBLP:conf/acl/ChengXHHWSL16} are shown to be effective. Multi-task learning is also examined for sequence labeling tasks~\cite{DBLP:conf/acl/Rei17,DBLP:conf/aaai/LiuSRXG0018}. In our paper, we study semi-supervised learning for keyphrase generation based on seq2seq models, which has not been explored before. Besides, we focus on leveraging source-side unlabeled articles to enhance performance with synthetic keyphrase construction or multi-task learning. 

\section{Neural Keyphrase Generation Model} 
\begin{figure}
\centering\includegraphics[width = 6.cm]{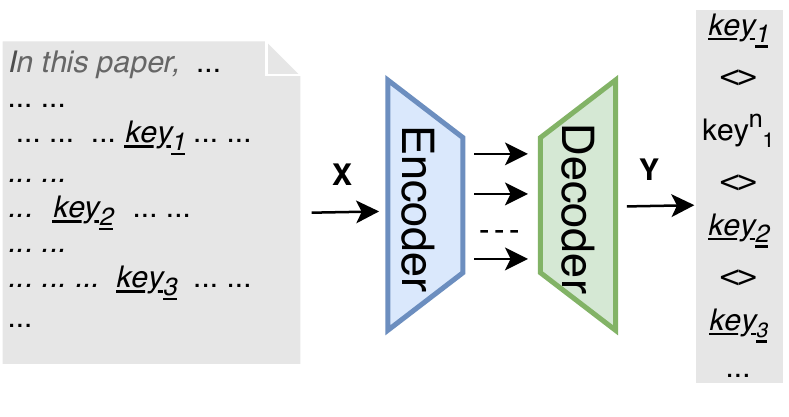}\caption{{Neural keyphrase generation model built on sequence-to-sequence framework. Input is the document, and output is the keyphrase sequence consisting of phrases present ($key_i$) or absent ($key^{n}_i$) in the input.}}
\label{model}
\end{figure}

In this section, we describe the neural keyphrase generation model built on a sequence-to-sequence model \cite{Seq2Seq} as illustrated in Figure \ref{model}. We denote the input source document as a sequence $\mathbf{x} = x_1 \cdots x_{|\mathbf{x}|}$ and its corresponding keyphrase set as $a = \{a_i\}_{i=1}^{|a|}$, with $a_i$ as one keyphrase.

\paragraph{Keyphrase Sequence Formulation.}
Different from the setup by \citet{DeepKey}, where input article is paired with each keyphrase $a_i$ to consist a training sample, 
we concatenate the keyphrases in $a$ into a keyphrase sequence $\mathbf{y} = a_1 \ \lozenge \  a_2 \ \lozenge \  \cdots \ \lozenge \ a_{|a|}$, where $\lozenge$ is a segmenter to separate the keyphrases\footnote{We concatenate keyphrases following the original keyphrase order in the corpora, and we set $\lozenge$ as ``;'' in our implementation. The effect of keyphrase ordering will be studied in the future work.}. With this setup, the seq2seq model is capable to generate all possible keyphrases in one sequence as well as capture the contextual information between the keyphrases from the same sequence. 

\paragraph{Seq2Seq Attentional Model.}
With source document $\mathbf{x}$ and its keyphrase sequence $\mathbf{y}$, an encoder encodes $\mathbf{x}$ into context vectors, from which a decoder then generates $\mathbf{y}$. 
We set the encoder as one-layer bi-directional LSTM model and the decoder as another one-layer LSTM model \cite{DBLP:journals/neco/HochreiterS97}. The probability of generating target sequence $p(\mathbf{y}|\mathbf{x})$ is formulated as: 
\begin{equation}
p(\mathbf{y}|\mathbf{x}) = \prod_{t=1}^{|\mathbf{y}|} p(y_t|\mathbf{y}_{<t}, \mathbf{x})
\label{overall_eq}
\end{equation}
where $\mathbf{y}_{<t} = y_1 \cdots y_{t-1}$.

Let $\mathbf{h}_t = [\overrightarrow{\mathbf{h}}_t; \overleftarrow{\mathbf{h}}_t]$ denote the hidden state vector in the encoder at time step $t$, which is the concatenation of forward hidden vector $\overrightarrow{\mathbf{h}}_t$ and backward hidden vector $\overleftarrow{\mathbf{h}}_t$. Specifically, $\overrightarrow{\mathbf{h}}_t = \mathbf{f}_{\text{LSTM}_{\text{e}}}(x_t, \overrightarrow{\mathbf{h}}_{t-1})$ and $\overleftarrow{\mathbf{h}}_t = \mathbf{f}_{\text{LSTM}_{\text{e}}}(x_t, \overleftarrow{\mathbf{h}}_{t+1})$, where $\mathbf{f}_{\text{LSTM}_{\text{e}}}$ is an LSTM unit in encoder. 

Decoder hidden state is calculated as $\mathbf{s}_t = \mathbf{f}_{\text{LSTM}_{\text{d}}}(y_{t-1}, \mathbf{s}_{t-1})$, where $\mathbf{f}_{\text{LSTM}_{\text{d}}}$ is an LSTM unit in decoder. We apply global attention mechanism \cite{DBLP:conf/emnlp/LuongPM15} to calculate the context vector:
\begin{align}
\nonumber \mathbf{c}_t = & \sum_{i=1}^{|\mathbf{x}|}\alpha_{t,i}\mathbf{h}_i  \\
\alpha_{t,i} = & \frac{ \exp(\mathbf{W}_{\text{att}}[\mathbf{s}_t;\mathbf{h}_i])}{\sum_{k=1}^{|x|}\exp(\mathbf{W}_{\text{att}}[\mathbf{s}_t \mathbf{h}_k])}
\end{align}
where $\alpha_{t,i}$ is the attention weight; 
$\mathbf{W}_{\text{att}}$ contains learnable parameters. In this paper, we omit the bias variables to save space. 

The probability to predict $y_t$ in the decoder at time step $t$ is factorized as:
\begin{align}
\nonumber p_{\text{vocab}}(y_t|\mathbf{y}_{<t}, \mathbf{c}_t) = \mathbf{f}_{\text{softmax}}(\mathbf{W}_{\text{d}_\text{1}} \cdot & \\ 
\tanh (&\mathbf{W}_{\text{d}_\text{2}}[\mathbf{s}_t;\mathbf{c}_t]) )
\end{align}
where $\mathbf{f}_{\text{softmax}}$ is the $\mathrm{softmax}$ function and $\mathbf{W}_{\text{d}_\text{1}}$ and $\mathbf{W}_{\text{d}_\text{2}}$ are learnable parameters.

\begin{algorithm}[t]
\caption{Keyphrase Ranking} \label{re-ranking}
\begin{algorithmic}[0]
\small
\Require Generated top $\mathrm{R}$ keyphrase sequences $\mathcal{S} = [\mathbf{y}_1, \cdots, \mathbf{y}_\mathrm{R}]$ ranked with generation possibility from high to low with beam search
\Ensure Ranked keyphrase set $\mathcal{A}$ with importance from high to low

\Function{Key-Rank}{$\mathcal{S}$}
\State{$\mathcal{A}$ $\gets$ \textcolor{blue}{$\mathrm{list}$}$\mathrm{()}$}\hfill{\AlgComment{$\text{\small{\emph{set $\mathcal{A}$ as a empty list.}}}$}}
\State{$\mathcal{Q} \gets \textcolor{blue}{\mathrm{dict}}()$}\hfill{\AlgComment{\small{\emph{to skip keyword that is already in $\mathcal{A}$}.}}}
\For{$\mathbf{y}_i$ $\in$ $\mathcal{S}$}{
\State{$y_i \gets \mathbf{y}_i.\textcolor{blue}{\mathrm{split}}(``\lozenge")$;}\hfill{\AlgComment{ \emph{split $\mathbf{y}_i$ by ``$\lozenge$''}}}
	\For{$a \in y_i$}{
    \State{\AlgComment{\emph{if $a$ has not been merged in $\mathcal{A}$, then keep it.}} }
    	\If{$\mathrm{not} \ \mathcal{Q}.\textcolor{blue}{\mathrm{has\_key}}(a)$}
		 \State{$\mathcal{A}.\textcolor{blue}{\mathrm{append}}(a)$}
           \State{ $\mathcal{Q}.\textcolor{blue}{\mathrm{update}}(\textcolor{blue}{\mathrm{dict}}(\{a:``"\}))$}
        \EndIf
    }\EndFor
}\EndFor
\EndFunction

\normalsize
\end{algorithmic}
\end{algorithm}

\paragraph{Pointer-generator Network.}
Similar to \citet{DeepKey}, we utilize copying mechanism via pointer-generator network \cite{pointer} to allow the decoder to directly copy words from input document, thus mitigating out-of-vocabulary (OOV) problem. 
At time step $t$, the generation probability $p_{\text{gen}}$ is calculated as:
\begin{equation}
p_{\text{gen}} = \mathbf{f}_{\text{sigmoid}}(\mathbf{W}_{\text{c}}\mathbf{c}_t + \mathbf{W}_{\text{s}}\mathbf{s}_t + \mathbf{W}_{\text{y}}y_t)
\end{equation}
where $\mathbf{f}_{\text{sigmoid}}$ is a $\mathrm{sigmoid}$ function; $\mathbf{W}_{\text{c}}$, $\mathbf{W}_{\text{s}}$ and $\mathbf{W}_{\text{y}}$ are learnable parameters. $p_{\text{gen}}$ plays a role of switcher to choose to generate a word from a fixed vocabulary with probability $p_{\text{vocab}}$ or directly copy a word from the source document with the attention distribution $\mathbf{\alpha}_t$. With a combination of a fixed vocabulary and the extended source document vocabulary, the probability to predict $y_t$ is:
\begin{equation}
p(y_t) = p_{\text{gen}} p_{\text{vocab}}(y_t|\mathbf{y}_{<t}, \mathbf{c}_t) + (1-p_{\text{gen}})\sum_{i:y_i=y_t}\alpha_{t,i}
\end{equation}
where if $y_t$ does not appear in the fixed vocabulary, then the first term will be zero; and if $y_t$ is outside source document, the second term will be zero. 


\begin{figure*}
\centering\includegraphics[width = \textwidth]{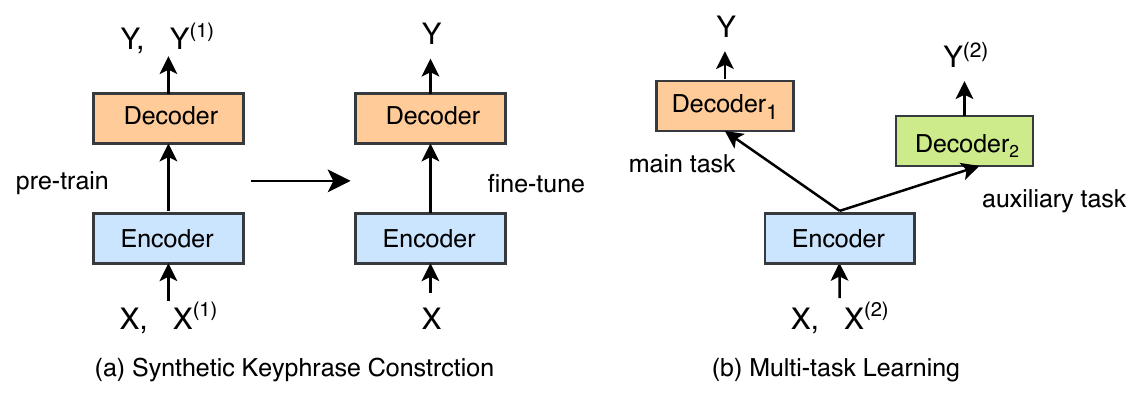}\caption{{Our two semi-supervised learning methods which are based on (a) synthetic keyphrase construction, and (b) multi-task learning. (X, Y) represents labeled sample. X$^{(1)}$ and X$^{(2)}$ denotes unlabeled documents. Y$^{(1)}$ refers to synthetic keyphrases of X$^{(1)}$ and Y$^{(2)}$ means the title of X$^{(2)}$. For method of synthetic keyphrase construction, model will be pre-trained on the mixture of gold-standards and synthetic data, then fine-tuned on labeled data. For multi-task learning, model parameters of main task and auxiliary task will be jointly updated. Encoder parameters of the two tasks are shared.} 
}
\label{training_procedure}
\end{figure*}

\paragraph{Supervised Learning.}
With a labeled dataset $\mathcal{D}_p = \{\mathbf{x}_{(i)}, \mathbf{y}_{(i)}\}_{i=1}^{N}$, the loss function of seq2seq model is as follows: 
\begin{equation}
\mathcal{L}(\theta) = -\sum_{i=1}^{N}\log p(\mathbf{y}_{(i)} | \mathbf{x}_{(i)};\theta)
\end{equation}
where $\theta$ contains all model parameters.

\paragraph{Keyphrase Inference and Ranking Strategy.}
Beam search is utilized for decoding, and the top $\mathrm{R}$ keyphrase sequences are leveraged for producing the final keyphrases. 
Here we use a beam size of $\mathrm{50}$, and $\mathrm{R}$ as $\mathrm{50}$. 
We propose a ranking strategy to collect the final set of keyphrases. Concretely, in sequence we collect unique keyphrases from the top ranked beams to lower ranked beams, and keyphrases in the same sequence are ordered as in the generation process. Intuitively, higher ranked sequences are likely of better quality. 
As for keyphrases in the same sequence, we find that more salient keyphrases are usually generated first. 
The ranking method is presented in Algorithm \ref{re-ranking}. 


\section{Semi-Supervised Learning for Keyphrase Generation}
As illustrated in Figure \ref{training_procedure}, two methods are proposed to leverage abundant unlabeled data. The first is to provide synthetic keyphrases using unsupervised keyphrase extraction methods or self-learning algorithm, then mixed with labeled data for model training, which is described in Section \ref{syn.key.gen}. Furthermore, we introduce multi-task learning that jointly generates keyphrases and the title of the document (Section \ref{sec:multi}). We denote the large-scale unlabeled documents as $\mathcal{D}_u =\{\mathbf{x}{'}_{(i)}\}_{i=1}^{M}$ and labeled data as $\mathcal{D}_p = \{\mathbf{x}_{(i)}, \mathbf{y}_{(i)}\}_{i=1}^{N}$, where $M \gg N$. 

\subsection{Synthetic Keyphrase Construction}\label{syn.key.gen}
The first proposed technique is to construct synthetic labeled data by assigning keyphrases for unlabeled documents, and then mix the synthetic data with human labeled data for modeling training. Intuitively, adding training samples with synthetic keyphrases has two potentially benefits: (1) the encoder is exposed to more documents in training, and (2) the decoder also benefit from additional information from identifying contextual information for keyphrases. 
We propose and compare two methods to extract synthetic keyphrases. 

\paragraph{Unsupervised Learning Methods.}
Unsupervised learning methods on keyphrase extraction have been long studied in previous work \cite{TextRank,SingleRank}. Here we select two effective and widely used methods to select keyphrases on unlabeled dataset $\mathcal{D}_u$, which are TF-IDF and TextRank \cite{TextRank}. We combine the two methods into a hybrid approach, in which we first adopt the two methods to separately select top $\mathrm{K}$ keyphrases from the document, we then take the union with duplicate removal. To construct the keyphrase sequence, we concatenate the terms from TF-IDF and then from TextRank, following the corresponding ranking order. 
We set $\mathrm{K}$ as $\mathrm{5}$ in our experiments.

\paragraph{Self-learning Algorithm.}
Inspired by prior work~\cite{NMT:Zhang,NMT:SennrichHB16}, we adopt self-learning algorithm to boost training data. 
Concretely, we first build a baseline model by training the seq2seq model on the labeled corpus $\mathcal{D}_p$. Then the trained baseline model is utilized to generate synthetic keyphrase sequence $\mathbf{y}'$ given a unlabeled document $\mathbf{x}'$. We adopt beam search to generate the synthetic keyphrase sequences and beam size is set as $\mathrm{10}$. The top one beam is selected. 

\paragraph{Training Procedure.}
After the synthetic data $\mathcal{D}_s = \{\mathbf{x}'_{(i)}, \mathbf{y}'_{(i)}\}_{i=1}^M$ is obtained by either of the aforementioned methods, we mix labeled data $\mathcal{D}_p$ with $\mathcal{D}_s$ to train the seq2seq model. 
As described in Algorithm \ref{method1}, we combine $\mathcal{D}_p$ with $\mathcal{D}_s$ into $\mathcal{D}_{p+s}$ and shuffle $\mathcal{D}_{p+s}$ randomly, then we pre-train the model on $\mathcal{D}_{p+s}$, in which no network parameters are frozen during the training process. 
The best performing model is selected based on validation set, then fine-tuned on $\mathcal{D}_{p}$ until converge. 

\begin{algorithm}[t]
\caption{Training Procedures for Synthetic
Keyphrase Construction} 
\label{method1}
\begin{algorithmic}[0]
\small
\Require $\mathcal{D}_p$, $\mathcal{D}_s$, $\theta$
\Ensure $\theta$

\Function{Pre-train}{$\mathcal{D}_p$, $\mathcal{D}_s$, $\theta$}
\State{$\mathcal{D}_{p+s} \gets \mathcal{D}_p \cup \mathcal{D}_s$}
\State{\textcolor{blue}{$\text{\it{Shuffle}}$} $\mathcal{D}_{p+s}$ \it{randomly} }
\State{\textcolor{blue}{$\text{\it{Update}}$} $\theta$ on $\mathcal{D}_{p+s}$ until converge}
\EndFunction

\Function{Fine-tune}{$\mathcal{D}_p$, $\theta$} 
\State{{\textcolor{blue}{\it{Set}} $\theta$ as the best parameters from \scshape{Pre-train}}}
\State{\textcolor{blue}{\it{Update}} $\theta$ on $\mathcal{D}_p$ until converge} 
\EndFunction
\normalsize
\end{algorithmic}
\end{algorithm}

\subsection{Multi-task Learning with Auxiliary Task}\label{sec:multi}
The second approach to leverage unlabeled documents is to employ a multi-task learning framework which combines the main task of keyphrase generation with an auxiliary task through parameter sharing strategy. 
Similar to the model structure in \citet{NMT:Zhang}, our main task and the auxiliary task share an encoder network but have different decoders. 
Multi-task learning will benefit from the source-side information to improve the model generality of encoder. 

In most domains such as scientific papers or news articles, a document usually contains a title that summarizes the core topics or ideas, with a similar spirit as keyphrases. We thus choose title generation as auxiliary task, which has been studied as a summarization problem~\cite{DBLP:conf/emnlp/RushCW15,DBLP:conf/naacl/ColmenaresLMS15}. 
Let $\mathcal{D}'_u = \{\mathbf{x}{'}_{(i)}, \mathbf{q}_{(i)}\}_{i=1}^M$ denote the dataset which is assigned with titles for unlabeled data $\mathcal{D}_u$, the loss function of multi-task learning is factorized as:
\begin{align}
\nonumber \mathcal{L}({\theta}^{{e}}, {\theta}^{{d}}_{\mathrm{1}}, \theta_{\mathrm{2}}^{{d}}) & = -\sum_{i=1}^N \log p({\mathbf{y}}_{(i)} | {\mathbf{x}}_{(i)};\theta^{{e}}, \theta^{{d}}_{\mathrm{1}})  \\
  & -\sum_{i=1}^M \log p({\mathbf{q}}_{(i)} | {\mathbf{x}}{'}_{(i)};\theta^{{e}}, \theta^{{d}}_{\mathrm{2}}) 
\end{align}
where $\theta^{{e}}$ indicates encoder parameters; $\theta^{{d}}_{\mathrm{1}}$ and $\theta^{{d}}_{\mathrm{2}}$ are the decoder parameters.

\paragraph{Training Procedure.}
We adopt a simple alternating training strategy to switch training between the main task and the auxiliary task. Specifically, we first estimate parameters on auxiliary task with $\mathcal{D}'_u$ for one epoch, then train model on the main task with $\mathcal{D}_p$ (labeled dataset) for $\mathrm{T}$ epochs. We follow this training procedure for several times until the model of our main task converges. We set $\mathrm{T}$ as $\mathrm{3}$. 

\section{Experiments}

\begin{table}
 \renewcommand{\arraystretch}{1.2}
\centering
 \fontsize{8.}{8.}\selectfont
\begin{tabular}{|l|c|c|c|}
\hline
\multicolumn{2}{|l|}{\textbf{Dataset}} & \scshape{\textbf{Train}} & \scshape{\textbf{Valid}} \\ \hline
\multicolumn{2}{|l|}{\underline{\emph{\textbf{Small-scale}}}} & \ & \ \\
\multicolumn{2}{|l|}{Document-Keyphrase} & $\mathrm{40,000}$ & $\mathrm{5,000}$ \\ 
\multicolumn{2}{|l|}{Document-SyntheticKeyphrase}  & $\mathrm{400,000}$ & N/A \\ 
\multicolumn{2}{|l|}{Document-Title} & $\mathrm{400,000}$ & $\mathrm{15,000}$ \\ 
\multicolumn{2}{|l|}{\underline{\emph{\textbf{Large-scale}}}} & \ & \ \\
\multicolumn{2}{|l|}{Document-Keyphrase} & $\mathrm{130,000}$ & $\mathrm{5,000}$ \\
\hline \hline
\textbf{Avg. \#Tokens in Train} & \scshape{\textbf{Labeled}} & \scshape{\textbf{Syn.}} & \scshape{\textbf{MTL}} \\ \hline
\underline{\emph{\textbf{Small-scale}}} & \ & \ & \ \\
Document & $\mathrm{176.3}$ & $\mathrm{175.9}$ & $\mathrm{165.5}$\\
Keyphrase Sequence & $\mathrm{23.3}$ & $\mathrm{23.5}$ & N/A \\
Title & N/A & N/A & $\mathrm{10.4}$ \\ \hline
\end{tabular}
\caption{Statistics of datasets used in our experiments.}
\label{dataset_statistic}
\end{table}

\begin{table*}[!htbp]
  \centering
  \fontsize{8.0}{8.}\selectfont
  \renewcommand{\arraystretch}{1.2}
  \begin{tabular}{|l | c|c | c|c | c|c | c|c | c|c|}
  \hline 
    \multirow{2}{*}{{\textbf{Model}}}  
    & \multicolumn{2}{c|}{\scshape{\textbf{Inspec}}}
    & \multicolumn{2}{c|}{\scshape{\textbf{Krapivin}}}
    & \multicolumn{2}{c|}{\scshape{\textbf{NUS}}}
    & \multicolumn{2}{c|}{\scshape{\textbf{SemEval}}}
    & \multicolumn{2}{c|}{\scshape{\textbf{KP20k}}}
    \\
    &  \textbf{\scshape{F$_1$@$\mathrm{5}$}}& \textbf{\scshape{F$_1$@$\mathrm{10}$}} & \textbf{\scshape{F$_1$@$\mathrm{5}$}}& \textbf{\scshape{F$_1$@$\mathrm{10}$}}
    &  \textbf{\scshape{F$_1$@$\mathrm{5}$}}& \textbf{\scshape{F$_1$@$\mathrm{10}$}} & \textbf{\scshape{F$_1$@$\mathrm{5}$}}& \textbf{\scshape{F$_1$@$\mathrm{10}$}} & \textbf{\scshape{F$_1$@$\mathrm{5}$}} & \textbf{\scshape{F$_1$@$\mathrm{10}$}} 
    \\  
    \hline 
    \underline{\emph{\textbf{Comparisons}}} & \ & \ & \ & \ & \ & \ & \ & \ & \ &
    \\
    \scshape{Tf-Idf}
    & $\mathrm{0.223}$ & $\mathrm{0.304}^\dagger$
    & $\mathrm{0.113}$ & $\mathrm{0.143}$
    & $\mathrm{0.139}$ & $\mathrm{0.181}$
    & $\mathrm{0.120}$ & $\mathrm{0.184}^\dagger$
    & $\mathrm{0.105}$ & $\mathrm{0.130}$
    \\ 
    \scshape{TextRank}
    & $\mathrm{0.229}$ & $\mathrm{0.275}$ 
    & $\mathrm{0.173}$ & $\mathrm{0.147}$
    & $\mathrm{0.195}$ & $\mathrm{0.190}$
    & $\mathrm{0.172}$ & $\mathrm{0.181}$
    & $\mathrm{0.181}$ & $\mathrm{0.150}$
    \\ 
	\scshape{SingleRank} 
    & $\mathrm{0.214}$ & $\mathrm{0.297}$
    & $\mathrm{0.096}$ & $\mathrm{0.137}$
    & $\mathrm{0.145}$ & $\mathrm{0.169}$
    & $\mathrm{0.132}$ & $\mathrm{0.169}$
    & $\mathrm{0.099}$ & $\mathrm{0.124}$
    \\
	\scshape{ExpandRank} 
    & $\mathrm{0.211}$ & $\mathrm{0.295}$
    & $\mathrm{0.096}$ & $\mathrm{0.136}$
    & $\mathrm{0.137}$ & $\mathrm{0.162}$
    & $\mathrm{0.135}$ & $\mathrm{0.163}$
    & N/A & N/A
    \\ 
      \scshape{Maui}
    & $\mathrm{0.041}$ & $\mathrm{0.033}$
    & $\mathrm{0.243}$ & $\mathrm{0.208}^\dagger$
    & $\mathrm{0.249}$ & $\mathrm{0.261}^\dagger$
    & $\mathrm{0.045}$ & $\mathrm{0.039}$
    & $\mathrm{0.265}$ & $\mathrm{0.227}^\dagger$
    \\  
	\scshape{KEA} 
    & $\mathrm{0.109}$ & $\mathrm{0.129}$
    & $\mathrm{0.120}$ & $\mathrm{0.131}$
    & $\mathrm{0.068}$ & $\mathrm{0.081}$
    & $\mathrm{0.027}$ & $\mathrm{0.027}$
    & $\mathrm{0.180}$ & $\mathrm{0.163}$
    \\  \hline \hline 

	\scshape{Seq2Seq-Copy}
    & $\mathrm{0.269}$ & $\mathrm{0.234}$
    & $\mathrm{0.274}$ & $\mathrm{0.207}$
    & $\mathrm{0.345}$ & $\mathrm{0.282}$
    & $\mathrm{0.278}$ & $\mathrm{0.226}$
    & $\mathrm{0.291}$ & $\mathrm{0.215}$
    \\  
    \underline{\emph{\textbf{Semi-supervised}}} & \ & \ & \ & \ & \ & \ & \ & \ & \ &
    \\
	\scshape{Syn.unsuper.} 
    & $\mathrm{\textbf{0.326}\ast}$ & $\mathrm{\textbf{0.334}\ast}$
    & $\mathrm{{0.283}}$ & $\mathrm{{0.239}\ast}$
    & $\mathrm{\textbf{0.356}}$ & $\mathrm{{0.317}\ast}$
    & $\mathrm{{0.306}}$ & $\mathrm{\textbf{0.294}\ast}$
    & $\mathrm{{0.300}\ast}$ & $\mathrm{\textbf{0.245}\ast}$
    \\  
    	\scshape{Syn.self-learn.} 
    & $\mathrm{{0.310}\ast}$ & $\mathrm{{0.301}\ast}$
    & $\mathrm{{0.289}}$ & $\mathrm{{0.236}\ast}$
    & $\mathrm{0.339}$ & $\mathrm{{0.305}}$
    & $\mathrm{{0.295}}$ & $\mathrm{{0.282}\ast}$
    & $\mathrm{{0.301}\ast}$ & $\mathrm{{0.240}\ast}$
    \\
    	\scshape{Multi-task} 
      & $\mathrm{\textbf{0.326}\ast}$ & $\mathrm{{0.309}\ast}$
    & $\mathrm{\textbf{0.296}}$ & $\mathrm{\textbf{0.240}\ast}$
    & $\mathrm{{0.354}}$ & $\mathrm{\textbf{0.320}\ast}$
    & $\mathrm{\textbf{0.322}}$ & $\mathrm{{0.289}\ast}$
    & $\mathrm{\textbf{0.308}\ast}$ & $\mathrm{{0.243}\ast}$
    \\
    \hline 
  \end{tabular}
  \caption{{Results of \emph{present} keyphrase generation with metrics F$_\mathrm{1}$@$\mathrm{5}$ and F$_\mathrm{1}$@$\mathrm{10}$. $\ast$ marks numbers that are significantly better than $\text{\scshape{Seq2seq-copy}}$ ($p<0.01$, $F$-test). Due to the high time perplexity, no result is reported by ExpandRank on \text{\scshape{KP20k}}, as done in \citet{DeepKey}.}}
   \label{tab:present}
\end{table*}

\subsection{Datasets}
Our major experiments are conducted on scientific articles which have been studied in previous work \cite{DBLP:conf/emnlp/Hulth03,DBLP:conf/icadl/NguyenK07,DeepKey}. We use the dataset from \citet{DeepKey} which is collected from various online digital libraries, e.g. ScienceDirect, ACM Digital Library, Wiley, and other portals. 

As indicated in Table \ref{dataset_statistic}, we construct a relatively small-scale labeled dataset with $\mathrm{40K}$ document-keyphrase\footnote{Here keyphrase refers to the keyphrase sequence.} pairs, and a large-scale dataset of $\mathrm{400K}$ unlabeled documents. Each document contains an abstract and a title of the paper. 
In multi-task learning setting, the auxiliary task is to generate the title from the abstract. 
For the validation set, we collect $\mathrm{5K}$ document-keyphrase pairs for the process of pre-training and fine-tuning for methods based on synthetic data construction. For multi-task learning, we use the same $\mathrm{5K}$ document-keyphrase pairs for the main task training, another $\mathrm{15K}$ document-title pairs for the auxiliary task. We further conduct experiments on a $\mathrm{130K}$ large-scale labeled dataset, which includes the small-scale labeled data.

Similar to \citet{DeepKey}, we test our model on five widely used public datasets from the scientific domain: \text{\scshape{Inspec}} \cite{DBLP:conf/emnlp/Hulth03}, \text{\scshape{NUS}} \cite{DBLP:conf/icadl/NguyenK07}, \text{\scshape{Krapivin}} \cite{krapivin2009large}, \text{\scshape{SemEval-2010}} \cite{DBLP:conf/semeval/KimMKB10} and \text{\scshape{KP20k}} \cite{DeepKey}. 

\subsection{Experimental Settings}
Data preprocessing is implemented as in \cite{DeepKey}. The texts are first tokenized by NLTK \cite{DBLP:conf/acl/BirdL04} and lowercased, then the numbers are replaced with $<$$\mathrm{digit}$$>$. 
We set maximal length of source text as $\mathrm{200}$, $\mathrm{40}$ for target text. Encoder and decoder both have a vocabulary size of $\mathrm{50K}$. 
The word embedding size is set to $\mathrm{128}$. 
Embeddings are randomly initialized and learned during training. 
The size of hidden vector is $\mathrm{512}$. 
Dropout rate is set as $\mathrm{0.3}$. Maximal gradient normalization is $\mathrm{2}$. Adagrad \cite{DBLP:journals/jmlr/DuchiHS11} is adopted to train the model with learning rate of $\mathrm{0.15}$ and the initial accumulator rate is $\mathrm{0.1}$.

For synthetic data construction, we first use batch size of $\mathrm{64}$ for model pre-training and then reduce to $\mathrm{32}$ for model fine-tuning. For both training stages, after $\mathrm{8}$ epochs, learning rate be decreased with a rate of $\mathrm{0.5}$. For multi-task learning, batch size is set to $\mathrm{32}$ and learning rate is reduced to half after $\mathrm{20}$ training epochs. To build baseline seq2seq model, we set batch size as $\mathrm{32}$ and decrease learning rate after $\mathrm{8}$ epochs. For self-learning algorithm, beam size is set to $\mathrm{10}$ to generate target sequences for unlabeled data and the top one is retained.

\subsection{Comparisons with Baselines}
\paragraph{Evaluation Metrics.} 
Following \cite{ DBLP:conf/conll/LiuCZS11,DeepKey}, we adopt top-N macro-averaged \emph{precision}, \emph{recall} and \emph{F-measure} ($F_1$) for model evaluation. \emph{Precision} means how many top-N extracted keywords are correct and \emph{recall} means how many target keyphrases are extracted in top-N candidates. Porter Stemmer is applied before comparisons. 

\paragraph{Baselines.}
We train a baseline seq2seq model on the small-scale labeled dataset. We further compare with four unsupervised learning methods: TF-IDF, TextRank \cite{TextRank}, SingleRank \cite{SingleRank}, ExpandRank \cite{SingleRank}, and two supervised learning methods of Maui \cite{Maui} and KEA \cite{KEA}. We follow \citet{DeepKey} for baselines setups.

\paragraph{Results.}
Here we show results for \emph{present} and \emph{absent} keyphrase generation separately\footnote{Recall that \emph{present} means the keyphrase appears in the document, otherwise, it is \emph{absent}.}. 
From the results of present keyphrase generation as shown in Table \ref{tab:present}, although trained on small-scale labeled corpora, our baseline seq2seq model with copying mechanism still achieves better F$_1$@$\mathrm{5}$ scores, compared to other baselines. 
This demonstrates that a baseline seq2seq model has learned the mapping patterns from source text to target keyphrases to some degree. However, small-scale labeled data still hinders the potential of seq2seq model according to the poor performance of F$_1$@$\mathrm{10}$. By leveraging large-scale unlabeled data, our semi-supervised learning methods achieve siginifcant improvement over seq2seq baseline, as well as exhibit the best performance in both F$_1$@$\mathrm{5}$ and F$_1$@$\mathrm{10}$ on almost all datasets. 

\begin{table}[t]
  \centering
  \fontsize{8.}{8.}\selectfont
  \renewcommand{\arraystretch}{1.2}
  \begin{tabular}{|l|c|c|c|c|}
    \hline
    & & \multicolumn{3}{r|}{\underline{\textbf{\emph{Semi-supervised}}}} \\
    \textbf{Dataset}
    & \scshape{Seq.} & \scshape{Syn.Un.} & \scshape{Syn.Self.} & \scshape{Multi.}
    \\    
    \hline
    \scshape{\textbf{Inspec}} & $\mathrm{0.012}$ & $\mathrm{{0.018}}$ & $\mathrm{{0.013}}$ & $\mathrm{\textbf{0.022}}$ \\    
    \scshape{\textbf{Krapivin}} & $\mathrm{0.01}$ & $\mathrm{0.008}$ & $\mathrm{{0.018}}$ & $\mathrm{\textbf{0.021}}$\\ 
    \scshape{\textbf{NUS}} & $\mathrm{0.002}$ & $\mathrm{{0.011}}$ & $\mathrm{{0.003}}$ & $\mathrm{\textbf{0.013}}$ \\     
    \scshape{\textbf{SemEval}} & $\mathrm{0.001}$ & $\mathrm{\textbf{0.008}}$ & $\mathrm{{0.003}}$ & $\mathrm{{0.006}}$\\   
    \scshape{\textbf{KP20k}} & $\mathrm{0.01}$ & $\mathrm{{0.016}}$ & $\mathrm{{0.013}}$ & $\mathrm{\textbf{0.021}}$\\  
    \hline
  \end{tabular} 
  \caption{{Results of \emph{absent} kephrase generation based on Recall@$\mathrm{10}$.}}
\label{absent_result}
\end{table}

We further compare seq2seq based models on the task of generating keyphrases beyond input article vocabulary. 
Illustrated by Table \ref{absent_result}, semi-supervised learning significantly improves the absent generation performance, compared to the baseline seq2seq. 
Among our models, the multi-task learning method is more effective at generating absent keyphrases than the two methods by leveraging synthetic data. The main reason may lie in that synthetic keyphrases potentially introduce noisy annotations, 
while the decoder in multi-task learning setting focuses on learning from gold-standard keyphrases. We can also see that the overall performances by all models are low, due to the intrinsic difficulty of absent keyphrase generation 
\cite{DeepKey}. Moreover, we only employ $\mathrm{40}$K labeled data for training, which is rather limited for training. Besides, we believe better evaluation methods should be used instead of exact match, e.g., by considering paraphrases. 
This will be studied in the future work.

\begin{figure}
\centering\includegraphics[width = \columnwidth]{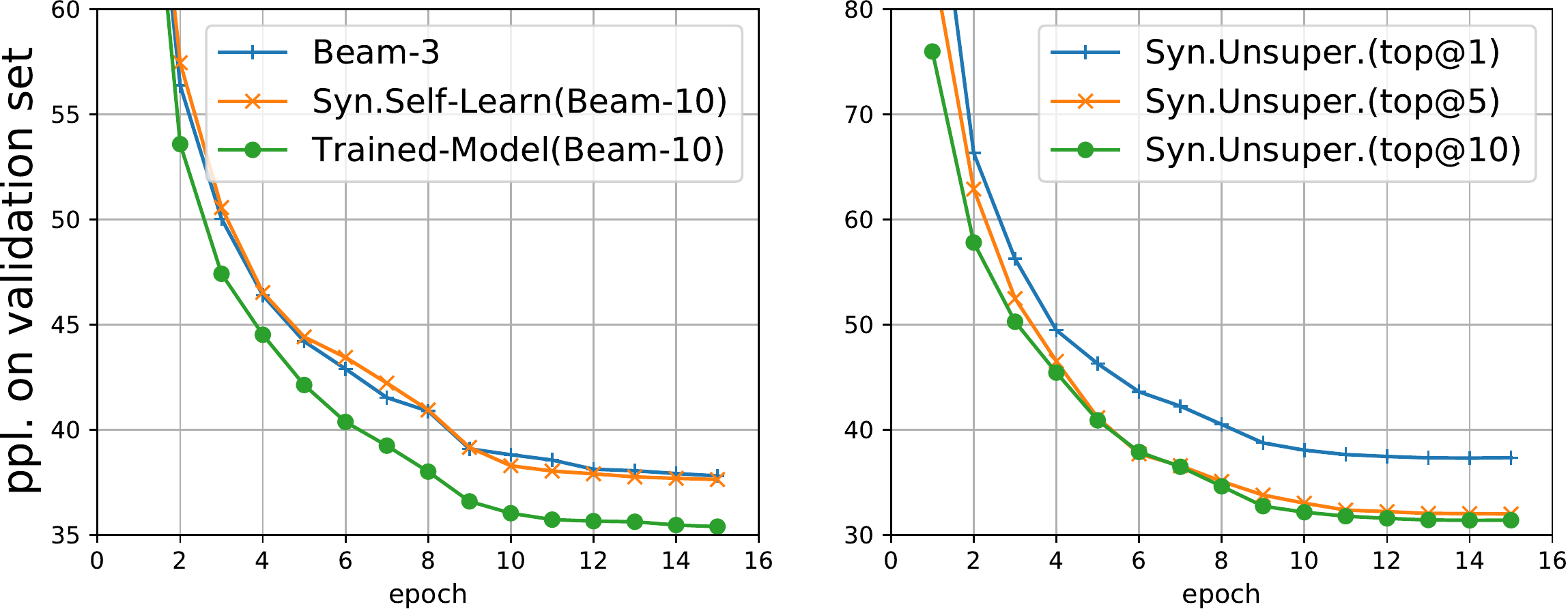}
\caption{{\emph{Pre-training} curves with perplexity on validation set with various options for synthetic keyphrases construction. Left is for options for self-learning algorithm and right is for unsupervised learning methods.}}
\label{data_quality_line}
\end{figure}

\subsection{Effect of Synthetic Keyphrase Quality}
In this section, we conduct experiments to further study the effect of synthetic keyphrase quality on model performance. Two sets of experiments are undertaken, one for evaluating unsupervised learning and one for self-learning algorithm. 

For self-learning algorithm, we further generate synthetic keyphrases using following options:
\begin{itemize}
\item Beam-size-3: Based on our baseline model trained with labeled data, we use beam search with a smaller beam size of $\mathrm{3}$ to generate synthetic data\footnote{We also experiment with greedy search (i.e. beam size of 1), however, unknown words are frequently generated.
}.
\item Trained-model: We adopt the model which has been trained with self-learning algorithm on $\mathrm{40K}$ labeled data and $\mathrm{400K}$ unlabeled data, to generate the top one keyphrase sequence with beam size of $\mathrm{10}$. 
\end{itemize}

For unsupervised learning method, we originally merge top-$\mathrm{K}$ ($\mathrm{K}$ = $\mathrm{5}$) keyphrases from TF-IDF and TextRank, here we use options where $\mathrm{K}$ is set as $\mathrm{1}$ or $\mathrm{10}$ to extract keyphrases: 

\begin{itemize}
\item Top@$\mathrm{1}$: Using TF-IDF or TextRank, we only keep top $\mathrm{1}$ extraction from each, then take the union of the two.

\item Top@$\mathrm{10}$: Similarly, we keep top $\mathrm{10}$ extracted terms from TF-IDF or TextRank, then take the union. 
\end{itemize}

\begin{figure}
\centering\includegraphics[width = \columnwidth]{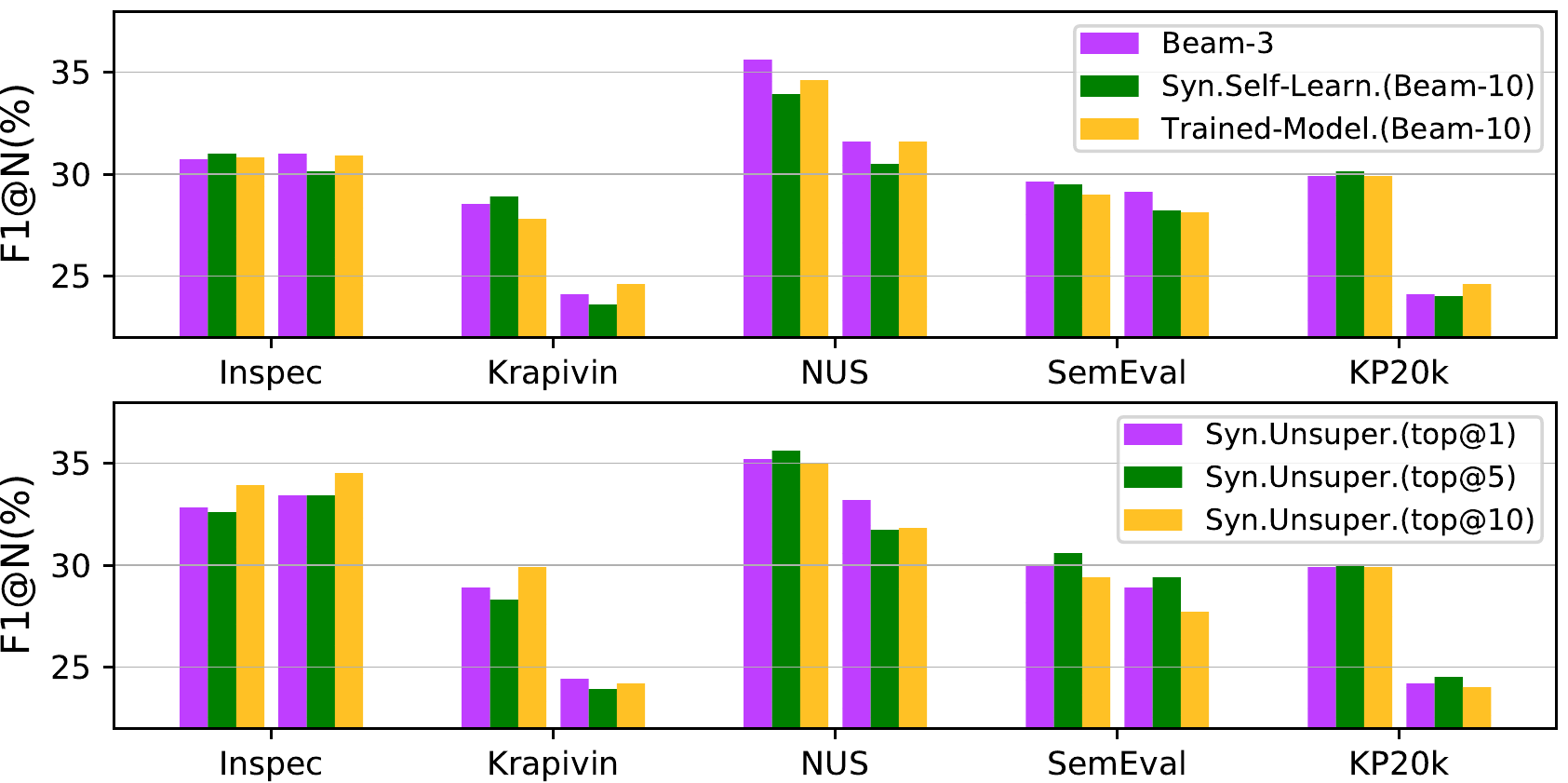}
\caption{{Effect of synthetic data quality on \emph{present} keyphrase generation (models are pre-trained and fine-tuned) based on F$_1$@$\mathrm{5}$ (left three columns) and F$_1$@$\mathrm{10}$ (right three columns), on five datasets. The upper is for self-learning algorithm and the bottom is for unsupervised learning method.}}
\label{data_quality_hist}
\end{figure}

As illustrated in Figure \ref{data_quality_line}, when models are pre-trained with synthetic keyphrases of better quality, results by ``Trained-model" consistently produce better performance (i.e., lower perplexity). 
Similar phenomenon can be observed when ``top@$\mathrm{5}$'' and ``top@$\mathrm{10}$'' are applied for extraction in unsupervised learning setting. 
%
Furthermore, after models are pre-trained and then fine-tuned, the results in Figure \ref{data_quality_hist} show that the difference among baselines becomes insignificant---the quality of synthetic keyphrases have limited effect on final scores. The reason might be that though synthetic keyphrases potentially introduce noisy information for decoder training, the encoder is still well trained. In addition, after fine-tuning on labeled data, the decoder acquires additional knowledge, thus leading to better performance and minimal difference among the options. 

\begin{figure}
\centering\includegraphics[width = \columnwidth]{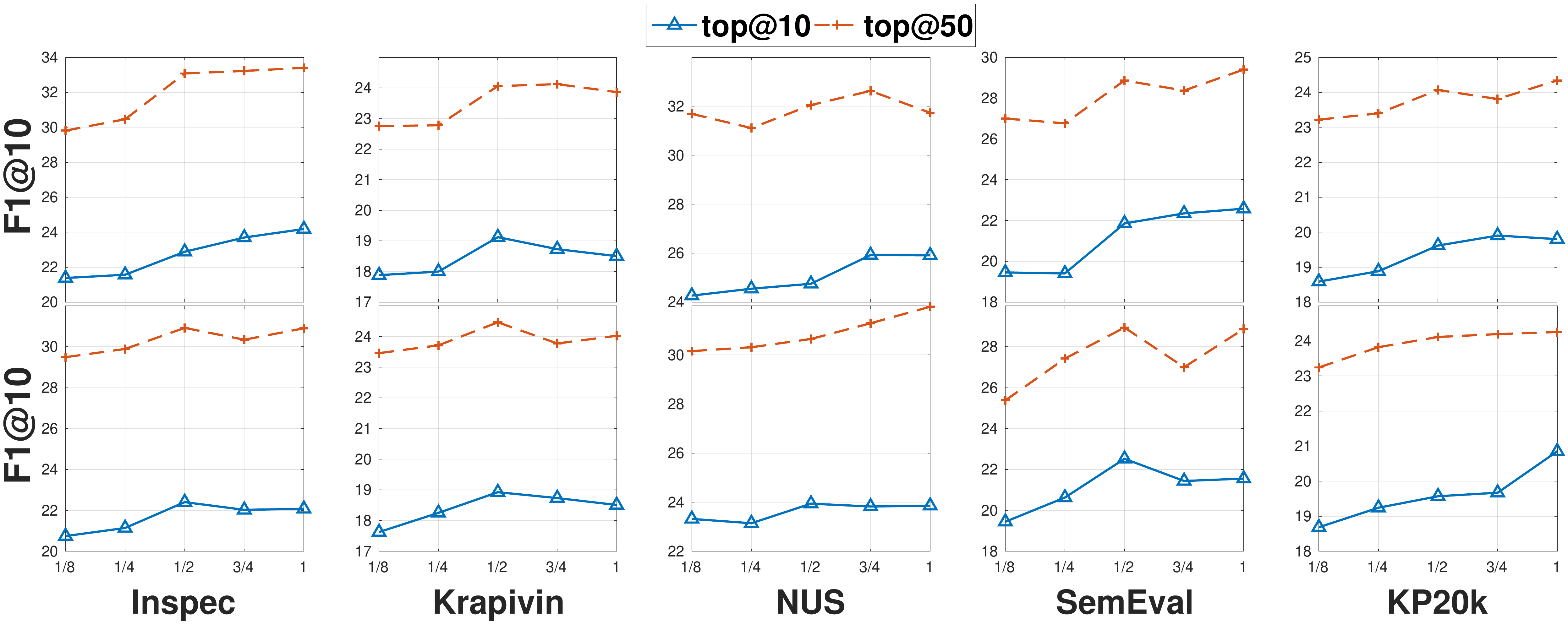}
\caption{{Effect of various amounts of unlabeled data for training on \emph{present} keyphrase generation with F$_1$@$\mathrm{10}$. Upper is for synthetic data construction method with unsupervised learning. Bottom is for multi-task learning algorithm.}}
\label{data_num_line}
\end{figure}

\begin{figure}
\centering\includegraphics[width = \columnwidth]{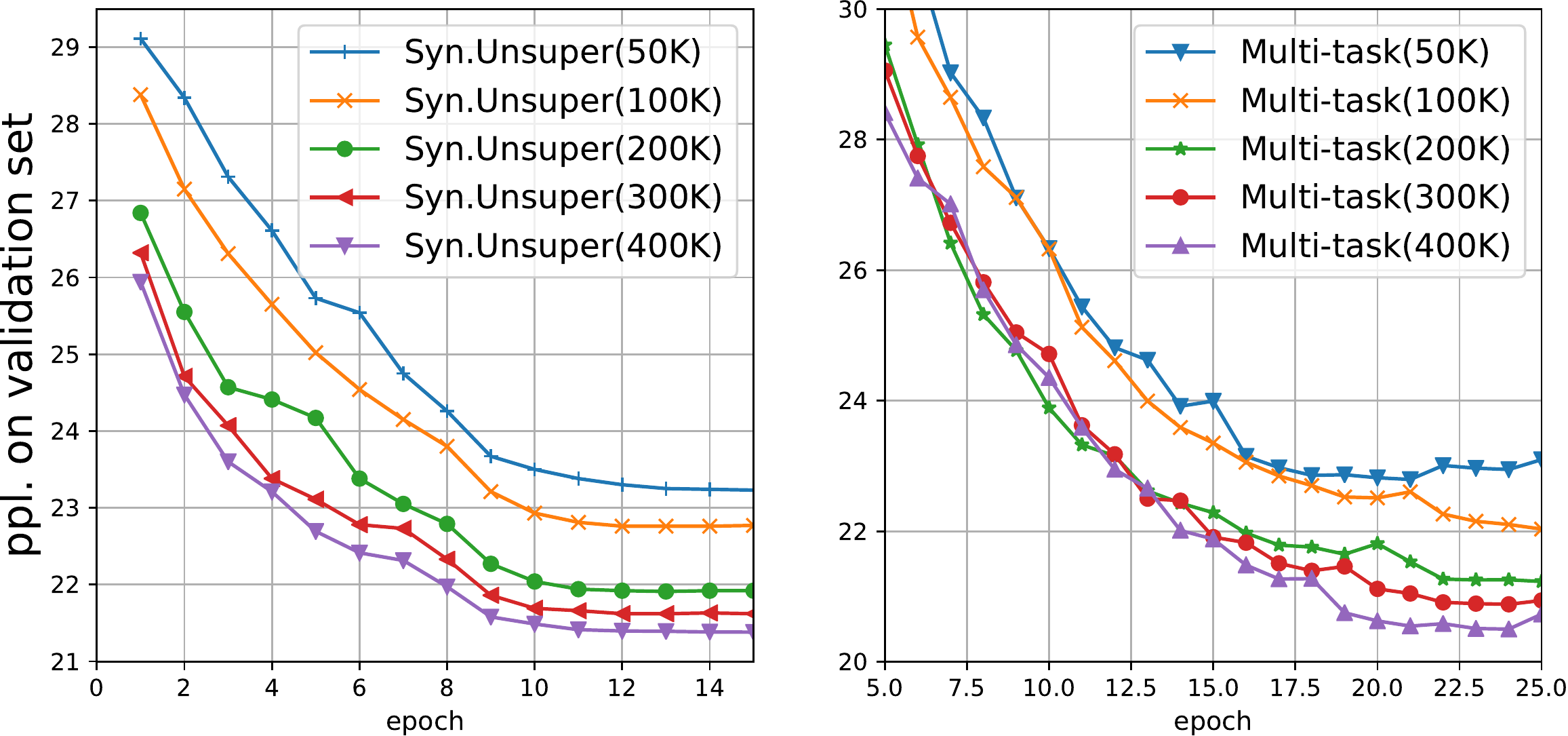}
\caption{{Perplexity on validation set with varying amounts of unlabeled data for training. Left is for fine-tuning procedure based on models trained with synthetic data constructed with unsupervised learning. Right is for multi-task learning procedure with performance on the main task.}}
\label{data_num_training}
\end{figure}

\subsection{Effect of Amount of Unlabeled Data}
In this section, we further evaluate whether varying the amount of unlabeled data will affect model performance. We conduct experiments based on methods of synthetic data construction with unsupervised learning and multi-task learning. We further carry out experiments with randomly selected $\mathrm{50K} (\mathrm{1/8})$, $\mathrm{100K} (\mathrm{1/4})$, $\mathrm{200K} (\mathrm{1/2})$ and $\mathrm{300K} (\mathrm{3/4})$ unlabeled documents from the pool of $\mathrm{400K}$ unlabeled data. After models being trained, we adopt beam search to generate keyphrase sequences with beam size of $\mathrm{50}$. We keep top N keyphrase sequences to yield the final keyphrases using Algorithm \ref{re-ranking}. F$_1$@$\mathrm{10}$ is adopted to illustrate the model performances. N is set as $\mathrm{10}$ or $\mathrm{50}$. 

The \emph{present} keyphrase generation results are shown in Figure \ref {data_num_line}, from which we can see that when increasing the amount of unlabeled data, model performance is further improved. This is because additional unlabeled data can provide with more evidence on linguistic or context features and thus make the model, especially the encoder, have better generalizability. This finding echoes with the training procedure illustrated in Figure \ref{data_num_training}, where more unlabeled data uniformly leads to better performance. 
Therefore, we believe that leveraging more unlabeled data for model training can boost model performance.

\begin{table}[!btbp]
  \centering
  \fontsize{8.}{8.}\selectfont
  \renewcommand{\arraystretch}{1.2}
  \begin{tabular}{|l|c|l|c|}
    \hline
    \textbf{Model}
    & \textbf{\scshape{F$_1$}} & \textbf{Model} & \textbf{\scshape{F$_1$}} 
    \\    
    \hline
    \underline{\emph{\textbf{Our Models}}} & \ & \underline{\emph{\textbf{Unsupervised}}} & \
    \\
    \scshape{Seq2Seq} & $\mathrm{0.056}$ & \scshape{TF-IDF} & $\mathrm{{0.270}}$ \\    
    \scshape{Syn.unsuper.} & $\mathrm{{0.083}}$ & \scshape{TextRank} & $\mathrm{{0.097}}$\\ 
    \scshape{Syn.self-learn.} & $\mathrm{{0.065}}$ & \scshape{SingleRank} & $\mathrm{{0.256}}$ \\     
    \scshape{Multi-task} & $\mathrm{\textbf{0.109}}$ & \scshape{ExpandRank} & $\mathrm{{0.269}}$ \\   
    \hline
  \end{tabular} 
  \caption{{Results of keyphrase generation for news from DUC dataset with F$_1$. Results of unsupervised learning methods are adopted from \citet{DBLP:conf/coling/HasanN10}.}}
\label{cross_domain}
\end{table}

\begin{table*}[!htbp]
  \centering
  \fontsize{8.}{8.}\selectfont
  \renewcommand{\arraystretch}{1.2}
  \begin{tabular}{|l| c|c| c|c| c|c| c|c| c|c|}
  \hline
  
    \multirow{2}{*}{\textbf{Model}}  
    & \multicolumn{2}{c|}{\scshape{\textbf{Inspec}}}
    & \multicolumn{2}{c|}{\scshape{\textbf{Krapivin}}}
    & \multicolumn{2}{c|}{\scshape{\textbf{NUS}}}
    & \multicolumn{2}{c|}{\scshape{\textbf{SemEval}}}
    & \multicolumn{2}{c|}{\scshape{\textbf{KP20k}}}
    \\
    &  \textbf{\scshape{F$_1$@$\mathrm{5}$}}& \textbf{\scshape{F$_1$@$\mathrm{10}$}} & \textbf{\scshape{F$_1$@$\mathrm{5}$}}& \textbf{\scshape{F$_1$@$\mathrm{10}$}}
    &  \textbf{\scshape{F$_1$@$\mathrm{5}$}}& \textbf{\scshape{F$_1$@$\mathrm{10}$}} & \textbf{\scshape{F$_1$@$\mathrm{5}$}}& \textbf{\scshape{F$_1$@$\mathrm{10}$}} & \textbf{\scshape{F$_1$@$\mathrm{5}$}} & \textbf{\scshape{F$_1$@$\mathrm{10}$}} 
    \\  
    \hline
    
    
    \scshape{Seq2Seq-Copy}
    & $\mathrm{\textbf{0.34}}$ & $\mathrm{0.329}$
    & $\mathrm{0.308}$ & $\mathrm{0.251}$
    & $\mathrm{0.36}$ & $\mathrm{0.327}$
    & $\mathrm{0.301}$ & $\mathrm{0.285}$
    & $\mathrm{0.318}$ & $\mathrm{0.251}$
    \\  
     \underline{\textbf{\emph{Semi-supervised}}} & \ & \ & \ & \ & \ & \ & \ & \ & \ &
    \\
	\scshape{Syn.unsuper.} 
    & $\mathrm{0.338}$ & $\mathrm{\textbf{0.34}}$
    & $\mathrm{{0.316}}$ & $\mathrm{\textbf{0.255}}$
    & $\mathrm{\textbf{0.365}}$ & $\mathrm{{0.335}}$
    & $\mathrm{\textbf{0.337}}$ & $\mathrm{{0.308}}$
    & $\mathrm{{0.322}}$ & $\mathrm{{0.261}\ast}$
    \\  
    	\scshape{Syn.self-learn.} 
    & $\mathrm{0.33}$ & $\mathrm{0.326}$
    & $\mathrm{0.304}$ & $\mathrm{\textbf{0.255}}$
    & $\mathrm{0.359}$ & $\mathrm{\textbf{0.336}}$
    & $\mathrm{{0.304}}$ & $\mathrm{{0.304}}$
    & $\mathrm{{0.321}}$ & $\mathrm{{0.263}\ast}$
    \\
    	\scshape{Multi-task} 
    & $\mathrm{0.328}$ & $\mathrm{0.318}$
    & $\mathrm{\textbf{0.323}}$ & $\mathrm{{0.254}}$
    & $\mathrm{\textbf{0.365}}$ & $\mathrm{0.326}$
    & $\mathrm{{0.319}}$ & $\mathrm{\textbf{0.312}}$
    & $\mathrm{\textbf{0.328}\ast}$ & $\mathrm{\textbf{0.264}\ast}$
    \\
    \hline
  \end{tabular}
  \caption{{Results of \emph{present} keyphrase generation on large-scale labeled data with F$_\mathrm{1}$@$\mathrm{5}$ and F$_\mathrm{1}$@$\mathrm{10}$. $\ast$ indicates significant better performance than $\text{\scshape{Seq2seq-copy}}$ with $p<0.01$ ($F$-test).}}
   \label{large_present_result}
\end{table*}

\subsection{A Pilot Study for Cross-Domain Test}
Up to now, we have demonstrated the effectiveness of leveraging unlabeled data for in-domain experiments, but is it still effective when being tested on a different domain? 
We thus carry out a pilot cross-domain test on news articles. The widely used DUC dataset~\cite{SingleRank} is utilized, consisting of $\mathrm{308}$ articles with $\mathrm{2,048}$ labeled keyphrases. 

The experimental results are shown in Table \ref{cross_domain} which indicate that: 1) though trained on scientific papers, our models still have the ability to generate keyphrases for news articles, illustrating that our models have learned some universal features between the two domains; and 2) semi-supervised learning by leveraging unlabeled data improves the generation performances more, indicating that our proposed method is reasonably effective when being tested on cross-domain data. 
Though unsupervised methods are still superior, for future work, we can leverage unlabeled out-of-domain corpora to improve cross-domain keyphrase generation performance, which could be a promising direction for domain adaption or transfer learning. 


\begin{table}[!htbp]
  \centering
  \fontsize{8.}{8.}\selectfont
  \renewcommand{\arraystretch}{1.2}
  \begin{tabular}{|l|c|c|c|c|}
    \hline
    &  & 
    \multicolumn{3}{r|}{\underline{\emph{\textbf{Semi-supervised}}}} \\ 
  {\textbf{Dataset}}  & {\scshape{Seq.}} & \scshape{Syn.Un.} & \scshape{Syn.Self.} & \scshape{Multi.}
    \\ \hline   
    \scshape{\textbf{Inspec}} & $\mathrm{0.021}$ & $\mathrm{{0.024}}$ & $\mathrm{{0.032}}$ & $\mathrm{\textbf{0.033}}$ \\    
    \scshape{\textbf{Krapivin}} & $\mathrm{0.02}$ & $\mathrm{{0.031}}$ & $\mathrm{{0.043}}$ & $\mathrm{\textbf{0.047}}$\\ 
    \scshape{\textbf{NUS}} & $\mathrm{0.009}$ & $\mathrm{{0.026}}$ & $\mathrm{{0.024}}$ & $\mathrm{\textbf{0.036}}$ \\     
    \scshape{\textbf{SemEval}} & $\mathrm{0.011}$ & $\mathrm{{0.014}}$ & $\mathrm{{0.015}}$ & $\mathrm{\textbf{0.02}}$\\   
    \scshape{\textbf{KP20k}} & $\mathrm{0.021}$ & $\mathrm{{0.034}}$ & $\mathrm{{0.039}}$ & $\mathrm{\textbf{0.046}}$\\  
    \hline
  \end{tabular} 
  \caption{{Performance on \emph{absent} keyphrase generation by Recall@$\mathrm{10}$ with large-scale labeled training data.}}
\label{large_absent_result}
\end{table}

\subsection{Training on Large-scale Labeled Data}
Finally, it would be interesting to study whether unlabeled data can still improve performance when the model is trained on a larger scaled labeled data. We conduct experiments on a larger labeled dataset with $\mathrm{130K}$ pairs, 
along with the $\mathrm{400K}$ unlabeled data. Here the baseline seq2seq model is built on the $\mathrm{130K}$ dataset. 

From the \emph{present} keyphrase generation results in Table \ref{large_present_result}, it can be seen that unlabeled data is still helpful for model training on a large-scale labeled dataset. 
This implies that we can also leverage unlabeled data to enhance generation performance even in a resource-rich setting. 
Referring to the \emph{absent} keyphrase generation results shown in Table \ref{large_absent_result}, semi-supervised learning also boosts the scores. From Table \ref{large_absent_result}, training on large-scale labeled data, \emph{absent} generation is significantly improved, compared to being trained on a small-scale labeled data (see Table \ref{absent_result}). 

\section{Conclusion and Future Work}
In this paper, we presented a semi-supervised learning framework that leverages unlabeled data for keyphrase generation built upon seq2seq models. We introduced synthetic keyphrases construction algorithm and multi-task learning to effectively leverage abundant unlabeled documents. 
Extensive experiments demonstrated the effectiveness of our methods, even in scenario where large-scale labeled data is available. 

For future work, we will 1) leverage unlabeled data to study domain adaptation or transfer learning for keyphrase generation; and 2) investigate novel models to improve \emph{absent} keyphrase generation when limited labeled data is available based on semi-supervised learning.

\section*{Acknowledgements}
This research is based upon work supported in part by National Science Foundation through Grants IIS-1566382 and IIS-1813341, and by the Office of the Director of National Intelligence (ODNI), Intelligence Advanced Research Projects Activity (IARPA), via contract \# FA8650-17-C-9116. The views and conclusions contained herein are those of the authors and should not be interpreted as necessarily representing the official policies, either expressed or implied, of ODNI, IARPA, or the U.S. Government. The U.S. Government is authorized to reproduce and distribute reprints for governmental purposes notwithstanding any copyright annotation therein. 
We thank three anonymous reviewers for their insightful suggestions on various aspects of this work.

\bibliographystyle{acl_natbib_nourl}
\bibliography{emnlp2018}

\end{document}